\documentclass[11pt]{article}

\usepackage[preprint]{acl}
\usepackage{graphicx}
\usepackage{times}
 \usepackage{booktabs} 
\usepackage{float}
\usepackage{amsmath}
\usepackage{amsfonts}
\usepackage[noabbrev,capitalize,nameinlink]{cleveref}
\usepackage{multirow}
\usepackage{latexsym}
\usepackage{wrapfig}

\usepackage[T1]{fontenc}

\usepackage[utf8]{inputenc}

\usepackage{microtype}

\usepackage{inconsolata}

\usepackage{graphicx}

\usepackage{tcolorbox}
\tcbuselibrary{breakable}

\usepackage[framemethod=TikZ]{mdframed}
\usepackage{xcolor}
\usepackage{fontawesome5}

\definecolor{black_kpmg}{HTML}{0e233e}
\hypersetup{
  urlcolor=blue,
}

\newcommand{\githubrepo}[2]{%
  \noindent
  {\faGithub\ }%
  \href{#2}{\texttt{#1}}%
}
\definecolor{black_kpmg}{HTML}{0e233e}

\definecolor{examplebg}{HTML}{FFFFFF}

\mdfdefinestyle{exampleboxstyle}{%
  backgroundcolor=examplebg,
  skipabove=\baselineskip,
  skipbelow=\baselineskip,
  linecolor=black_kpmg,
  linewidth=2pt,
  roundcorner=5pt,                      
  innertopmargin=1.5em,                  
  innerbottommargin=1.5em,
  innerleftmargin=1.5em,
  innerrightmargin=1.5em,
  frametitleaboveskip=4pt,
  frametitlebelowskip=4pt,
  frametitlealignment=\raggedright,     
  frametitlebackgroundcolor=black_kpmg,   
  frametitlefont=\color{white}\bfseries, 
}
\newenvironment{examplebox}[1]{%
  \begin{mdframed}[style=exampleboxstyle, frametitle={#1}]%
}{
  \end{mdframed}
}
\title{Probing for Knowledge Attribution in Large Language Models}

\author{
  Ivo Brink \\
  KPMG NL / University of Amsterdam\\
  Amsterdam, The Netherlands \\
  \texttt{ivobrink00@gmail.com}
  \And
  Alexander Boer \\
  KPMG NL \\
  Amsterdam, The Netherlands \\
  \texttt{boer.alexander@kpmg.nl}
  \And
  Dennis Ulmer \\
  University of Amsterdam \\
  Amsterdam, The Netherlands \\
  \texttt{dennis.ulmer@mailbox.org}
}

\begin{document}
\maketitle
\begin{abstract}
Large language model (LLM) hallucinations, meaning fluent but factually incorrect generations, fall into two types: faithfulness violations, where the model misuses provided context, and factuality violations, where answers reflect errors in internal knowledge. Proper mitigation depends on knowing which source drives each answer. We study \emph{contributive attribution}, i.e.\@ the classification of the dominant knowledge source behind each output, and show that a simple linear probe trained on hidden representations can reliably identify it. We introduce \textsc{AttriWiki}, a self-supervised pipeline that automatically generates labeled training data by prompting models to recall withheld entities from memory or read them from context without relying on knowledge conflicts. Probes trained on \textsc{AttriWiki} achieve up to 0.96 Macro-$F_1$ on Llama-3.1-8B, Mistral-7B, and Qwen2.5-7B, transfer to SQuAD and WebQuestions with 0.92--0.99 accuracy, and generalize zero-shot to \citet{tighidet2024probinglanguagemodelsknowledge}'s benchmark, outperforming their probe on conflicting settings without retraining. Furthermore, attribution mismatches raise error rates by up to 70\%, though correct attribution does not guarantee correct answers, pointing to the need for broader detection frameworks. \\
\githubrepo{IvoBrink/AttriWiki}{https://github.com/IvoBrink/RACDH}
\end{abstract}

\section{Introduction}
Picture yourself rushing to book a last-minute flight for a funeral. You ask an airline’s chatbot whether you can claim a bereavement fare, and the system confidently replies, ``Yes, you have 90 days.'' The rule, however, never existed, and months later, the refund is denied. This example stems from a real case in which Air Canada was held liable for its chatbot’s misinformation \citep{cecco2024air, lifshitz2024bctribunal}. A single inaccurate response can now impose serious financial, legal, or practical consequences.

Although LLM hallucinations, untrue or ungrounded outputs \citep{survey_taxonomy_hallucinations}, have become a central research focus, existing detection techniques often miss a more fundamental issue: users lack transparency over where a model’s answers come from. 
The airline could have benefited from knowing the chatbot’s true provenance, as attribution would have enabled verification against official policy. Prior approaches, including LLM-as-a-judge \citep{manakul2023selfcheckgptzeroresourceblackboxhallucination, ravi2024lynxopensourcehallucination, friel2023chainpollhighefficacymethod} and uncertainty measures \citep{tomani2024uncertainty, shelmanov2025head}, struggle because models can be confidently wrong \citep{kuhn2023semanticuncertaintylinguisticinvariances, simhi2025trust} and because ``hallucination'' lacks a consistent operational definition \citep{qi2024catchelephantsurveyevolvement}.  

To address this gap, we turn to \emph{contributive attribution} \citep{worledge2023unifyingcorroborativecontributiveattributions}, which distinguishes between contextual knowledge (from the prompt or retrieved evidence) and parametric knowledge (stored in the model’s weights). This framing complements the well-known divide between faithfulness and factuality errors \citep{survey_taxonomy_hallucinations, qi2024catchelephantsurveyevolvement, ye2023cognitivemiragereviewhallucinations} within hallucination taxonomies, but shifts the focus: instead of asking whether the output is true, we ask whether the model relied on the source the user intended. Attribution signals can help users detect when a model defaults to conflicting parametric knowledge or, conversely, when it overly relies on irrelevant context. This is particularly valuable in retrieval-augmented generation systems \citep{Lewis2020RetrievalAugmentedGF, RAGpretrain, RAGTrillionTokens, AtlasRAG}, where errors often arise not from bad retrieval but from the model ignoring retrieved information \citep{petroni2020how, li-etal-2023-large}. 

Our contributions are as follows:
Using the \textsc{AttriWiki} pipeline involving Wikipedia passages, entity selection, and controlled prompt variations, we create a dataset in which each completion has a clear, verifiable knowledge source (parametric or contextual). During generation, we record hidden states at key token positions, producing attribution features without manual labels or conflicting sources. We then train lightweight probing classifiers on these representations to distinguish contextual from parametric retrieval and test their generalization on external QA datasets such as \textsc{WebQuestions} and \textsc{SQuAD} \citep{berant-etal-2013-semantic, rajpurkar2016squad100000questionsmachine} and compare to existing attribution probing work \citep{tighidet2024probinglanguagemodelsknowledge}. We show that contributive attribution is linearly decodable from LLM hidden states, particularly from the middle to upper layers. Attribution mismatches, in which a model answers from the wrong knowledge source, significantly increase error rates, especially in misleading contexts. Yet, correct attribution alone does not guarantee factual correctness.
\textsc{AttriWiki} and all code  are publicly available.\footnote{The repository is available under \url{https://github.com/IvoBrink/AttriWiki}}

\section{Related Work}
\paragraph{Knowledge in LLMs.}
Understanding hallucinations ultimately requires understanding how large language models acquire, store, and retrieve knowledge. Factual knowledge in LLMs is parametric and emergent: no single parameter encodes a fact, but distributed activation patterns collectively give rise to factual recall \citep{dai2022knowledge, yao2024knowledge, wang2025functional}. 
Early probing work, such as LAMA \citep{petroni2019languagemodelsknowledgebases}, showed that factual recall in language models is brittle, varying substantially across languages, paraphrases, and prompt formulations \citep{kassner-etal-2021-multilingual, elazar2021measuring, jiang-etal-2020-know}.
In addition, recent works formalize knowledge in models by mapping philosophical notions of knowledge onto model behavior: Under the ``true-belief'' framework \citep{fierro2024definingknowledgebridgingepistemology} for instance, a fact is known by a model if it is reproduced consistently, which we use to make a practical distinction between parametric knowledge and context-derived information.

\paragraph{Attribution and Knowledge Conflicts.}
Once parametric and contextual knowledge are distinguished, a central challenge is detecting their usage. Knowledge conflicts arise when parametric beliefs contradict retrieved or provided context, a situation common in retrieval-augmented settings \citep{petroni2020how, li-etal-2023-large}. Attribution methods aim to trace generated content back to its source. Retrieval-augmented generation approaches \citep{Lewis2020RetrievalAugmentedGF, AtlasRAG} expose evidence to the model, but cannot guarantee its use, and models frequently hallucinate citations even when retrieval is correct \citep{zuccon2023chatgpthallucinatesattributinganswers}.  
Model-based attribution methods target contributive attribution through architectural modifications that encode source identifiers \citep{khalifa2024sourceawaretrainingenablesknowledge}, probing approaches that infer whether outputs derive from memory or context \citep{tighidet2024probinglanguagemodelsknowledge}, or component-level analyses that localize the attention heads responsible for each retrieval mode \citep{atlascontext}. The first requires substantial architectural changes, while the latter two both depend on counterfactual prompts in which context contradicts parametric knowledge, and therefore conflate provenance with conflict resolution.
Recent mechanistic work has begun to clarify how conflicting knowledge signals are processed. \citet{zhao2025understanding} trace entity-level knowledge flows through the model, showing that parametric and contextual knowledge are routed through largely distinct attention circuits and coexist as superposed signals rather than competing via direct suppression. Their entity-conditioned probes analyze how competing signals evolve across a forward pass, making their setup better suited to study conflict resolution dynamics rather than genuine source detection in the absence of conflict. Nonetheless, this perspective provides a mechanistic foundation for attribution: because both knowledge sources are routed through largely distinct circuits and persist as superposed signals, their relative contribution might remain detectable even without an explicit conflict. 

\paragraph{Probing LLM Representations.}
Probing hidden states offers a lightweight approach for studying knowledge usage directly from model internals. This line of work builds on a long tradition of analyzing how linguistic and factual information is encoded across transformer layers \citep{conneau2018cramsinglevectorprobing, tenney2019learncontextprobingsentence, hewitt-manning-2019-structural}. Prior studies show that layers specialize in surface, syntactic, and semantic features \citep{jawahar-etal-2019-bert}, that factual associations emerge in neuron-level key–value structures \citep{geva-etal-2021-transformer}, and that hidden states reflect latent capabilities such as lying \citep{azaria2023internalstatellmknows}, causal reasoning \citep{rohekar2024causalworldrepresentationgpt}, and instruction following \citep{heo2024llmsknowinternallyfollow}. These findings suggest that signals relevant to attribution and provenance are also present in model activations.  
Probing for factual knowledge has already been explored by a series of works (see \citealp{youssef2023give} for an overview), but recent work applies probing specifically to hallucination detection. MIND \citep{su2024unsupervisedrealtimehallucinationdetection} trains classifiers on final-layer activations, outperforming SelfCheckGPT \citep{manakul2023selfcheckgptzeroresourceblackboxhallucination} and GPT-4o critics on several benchmarks. However, it provides limited interpretability regarding why a response is classified as hallucinated.

Overall, prior work offers strong taxonomies and attribution techniques, but a key gap remains: existing methods lack scalable, controlled datasets that cleanly isolate parametric versus contextual knowledge use. This motivates approaches that explicitly construct such contrasts and probe hidden states to reveal the true origin of model outputs.

\section{A Self-Supervised Data Pipeline}

\subsection{Motivation}
Although recent work examines the use of parametric–contextual knowledge, existing datasets do not isolate retrieval modes. Many studies expose both sources simultaneously through conflict settings and infer attribution \emph{post hoc} from the model’s answer \citep{tighidet2024probinglanguagemodelsknowledge, xu2024knowledgeconflictsllmssurvey, zhao2025understanding}, conflating provenance with downstream decision-making and conflict resolution.

Therefore, checking whether a fact is present in a model’s parametric memory is necessary to isolate contextual vs.\@ parametric examples. Prior work verifies parametric knowledge via data recency \citep{zhang2024evaluatingexternalparametricknowledge} or by querying the model \citep{cheng2024understandinginterplayparametriccontextual}. However, these approaches neither explicitly detect attribution nor prevent the simultaneous availability of parametric and contextual knowledge when analyzing their interactions.

To this end, we introduce \textsc{AttriWiki}, a self-supervised data pipeline that generates paired examples which \emph{force} models to draw information either from context or from stored knowledge, established through extensive knowledge testing. By ensuring that only a single knowledge source is available at generation time, \textsc{AttriWiki} enables isolated attribution analysis without manual labeling.

\begin{figure}[t]
\centering
\includegraphics[width=\linewidth,height=0.55\textheight,keepaspectratio]{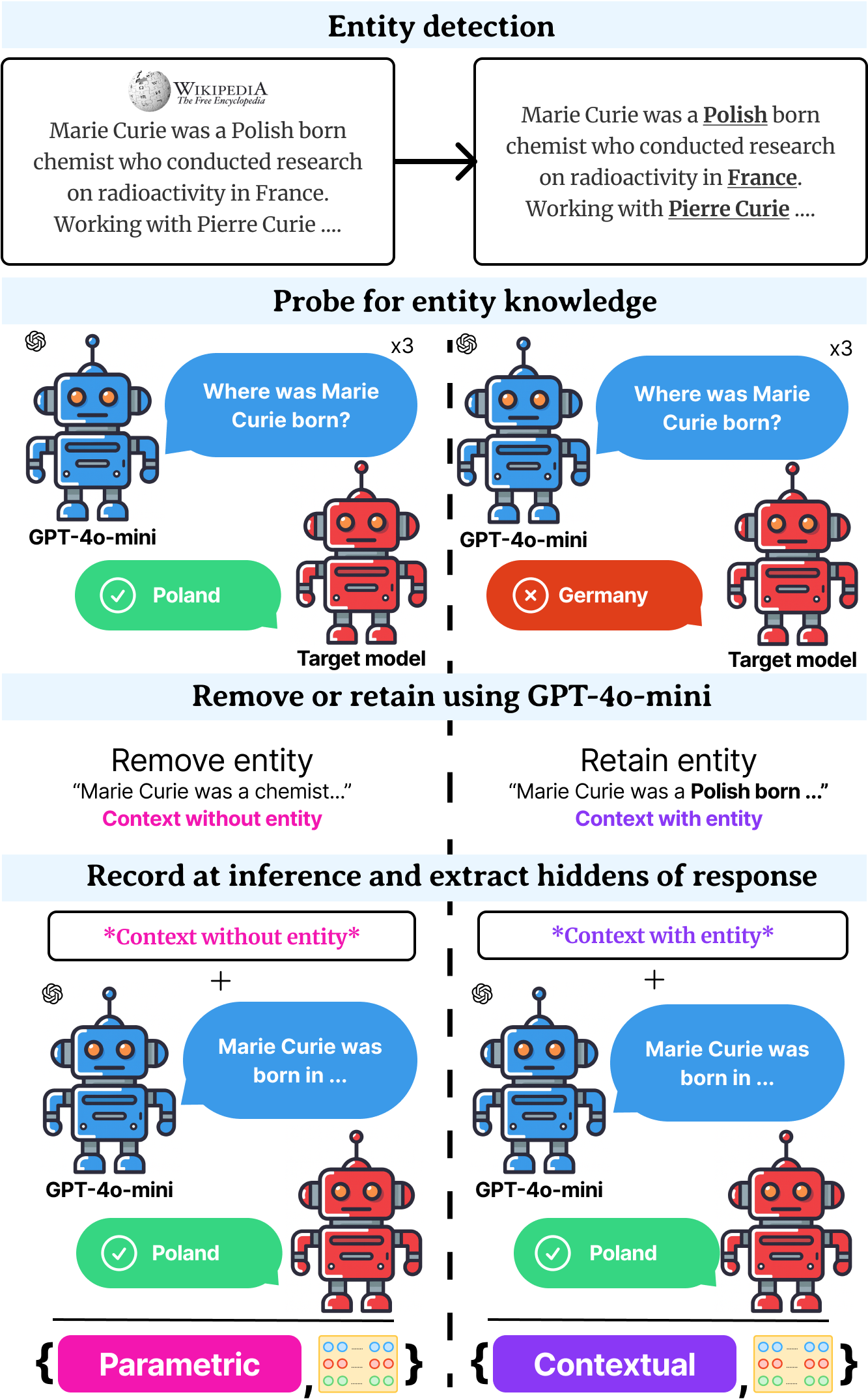}
\caption{Overview of the data generation pipeline, showing entity selection, knowledge testing, prompt construction, and hidden-state extraction.}
\label{fig:pipeline}
\end{figure}

\subsection{Data and Models}
\paragraph{Data.}
Wikipedia is part of many modern pretraining corpora \citep{jiang2023mistral7b, grattafiori2024llama3herdmodels, qwen2025qwen25technicalreport}. 
We use its dense factual text to test both retrieval modes naturally.
Our knowledge-testing procedure (\cref{section3:method}) supports the claim that mid-sized open language models (approximately 7--8B parameters) only partially memorize Wikipedia facts: roughly half of sampled entities are recalled through our knowledge tests and the rest are not (\cref{tab:model_stats}), indicating that neither retrieval mode is degenerate for any target model. 
We sample 20k pages using the MediaWiki API,\footnote{See \url{https://en.wikipedia.org/w/api.php} (last accessed May 20, 2026). The Wikipedia text is licensed under \textbf{CC BY-SA 3.0/4.0}, and all use complies with the terms of that license.} discarding snippets shorter than 300 characters.
\paragraph{Data Generation Model.}
We use \textsc{GPT-4o-mini} (\texttt{gpt-4o-mini}; \citealp{openai2024gpt4omini}) as an instruction-following model to paraphrase passages, remove or retain entities, and rewrite completion sentences during dataset construction (see next section).

\paragraph{Target Models.}
Attribution probing is performed on open-weight language models from which we extract hidden states during generation. We use \textsc{Llama-3.1-8B} \citep{grattafiori2024llama3herdmodels}, \textsc{Mistral-7B} \citep{jiang2023mistral7b}, and \textsc{Qwen2.5-7B} \citep{qwen2025qwen25technicalreport} as target models.

\subsection{Method}
\label{section3:method}
An overview of the data generation pipeline is shown in \cref{fig:pipeline}. Because attribution concerns the source of specific factual content, we focus on \emph{named entities} as attribution anchors, identified using spaCy’s transformer-based NER \citep{honnibal2020spacy}. Entities that overlap, are synonymous, or closely resemble the passage title are found using a RoBERTa \citep{liu2019robertarobustlyoptimizedbert} cross-encoder (CE) similarity model.\footnote{ \texttt{cross-encoder/stsb-roberta-base}.} The CE assigns pairwise semantic similarity scores, and entities above a threshold of $0.6$ are considered synonyms (we refer to this procedure as \emph{synonym matching}).\footnote{The threshold of 0.6 is anchored to the cross-encoder's training scale: it is trained on STS-B, whose 0–5 human ratings are rescaled to [0,1] \citep{cer-etal-2017-semeval}, so 0.6 corresponds to a rating of 3 (roughly equivalent). The value sets how strictly two strings must correspond to denote the same entity, controlling both the permissiveness of our knowledge proxy in accepting similar recalled completions and the similarity a candidate entity may bear to the passage title before exclusion.} In this case, synonyms are excluded. From the remaining entity candidates, \textsc{GPT-4o-mini} selects up to three representative, non-numeric, non-temporal entities per passage (see \cref{app:entity_selection}).

Each entity--passage pair undergoes \emph{knowledge testing} to determine whether the target model can recall the entity without the passage, serving as an \emph{operational proxy} for parametric memory. We use three prompt formats to test entity knowledge (see \cref{app:knowledge}).
If any test elicits the correct entity (exact or \emph{synonym match}), it is labeled \emph{known}; otherwise, \emph{unknown}. This high-recall labeling avoids false unknown labels, preferring discarding an example over admitting one that could draw on both sources.

We then construct \emph{contextual} and \emph{parametric} variants for each passage. For known entities, all mentions are removed using \textsc{GPT-4o-mini} (\cref{app:entity_removal}), thereby forcing parametric retrieval; for unknown entities, the name remains visible, while a different entity is removed to prevent the lexical bias this may introduce. \textsc{GPT-4o-mini} appends a short, natural completion cue such as ``The [description] is called~…,'' yielding paired prompts that differ only in the presence of the entity (see \cref{app:appending_sentence}).

Each prompt is submitted to the target model to produce a completion. During greedy decoding, we extract hidden representations at two key token positions: (i) the first generated token (FTG), which reflects the model’s initial response intent allowing us to compare responses without specific entities \citep{su2024unsupervisedrealtimehallucinationdetection}, and (ii) the last token of the entity (LTE), capturing the fully integrated entity semantics \citep{tighidet2024probinglanguagemodelsknowledge}.\footnote{We omit first-token-entity as it coincides with first-token-generation in almost all cases.} Entity spans are located by exact or \emph{synonym matching}, and the corresponding hidden states are serialized for attribution classification. This automated pipeline produces large volumes of data, with each example having a verifiable knowledge source (see \cref{app:resources} for an overview of the resource consumption).

\begin{table}[tb]
\centering
\renewcommand*{\arraystretch}{1.1}
\scalebox{0.95}{
\begin{tabular}{lrrr}
\toprule
\textbf{\textsc{AttriWiki}} & \textbf{Ratio (P:C)} & \textbf{Syn} & \textbf{Size} \\
\midrule
Llama-3.1-8B    & $\approx$5{:}3 & 10.3\% & 27{,}244 \\
Mistral-7B-v0.1 & $\approx$3{:}2 & 10.7\% & 26{,}853 \\
Qwen2.5-7B      & $\approx$1{:}1 & 12.0\% & 24{,}336 \\
\bottomrule
\end{tabular}
}
\caption{\textsc{AttriWiki} dataset composition for each model. \textbf{Ratio (P:C)} denotes the approximate parametric-to-contextual proportion of examples. \textbf{Syn} denotes the synonym match rate, and \textbf{Size} denotes the total number of examples.}
\label{tab:model_stats}
\end{table}

\paragraph{Results.} \cref{tab:model_stats} summarizes the \textsc{AttriWiki} dataset for Llama-3.1-8B, Mistral-7B-v0.1, and Qwen2.5-7B. Llama and Mistral show parametric-dominant distributions ($\approx$5{:}3 and 3{:}2), while Qwen is nearly balanced ($\approx$1{:}1). Dataset sizes are similar across models.  Synonym match rates (percentage of completions that matched semantically but not verbatim) are highest for Qwen (12.0\%). Overall, the pipeline behaves consistently.

\begin{figure*}[tb]
    \centering
    \includegraphics[width=\linewidth]{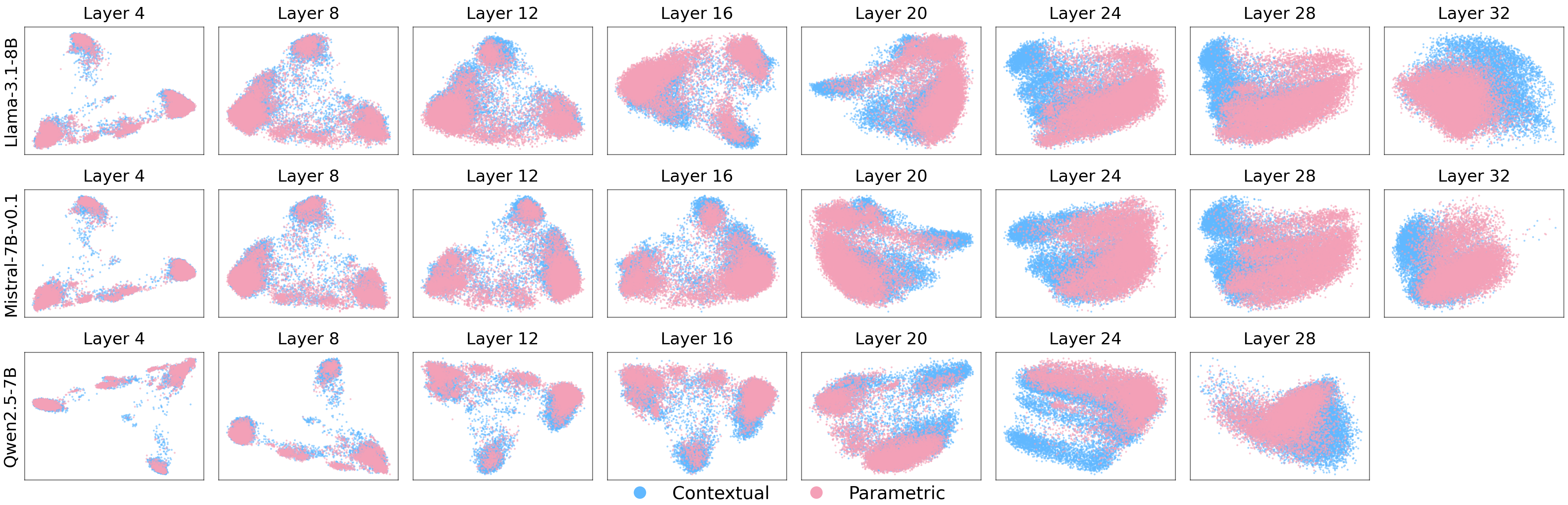}
    \caption{Per-layer PCA of decoder hidden states at the first generated token (FTG). Qwen has 28 layers (vs. 32 in Llama and Mistral). In the mid to upper layers, contextual and parametric activations increasingly diverge, indicating greater separability. Plots for all layers are given in \cref{app:latent-space}.}
    \label{fig:PCA_activations}
\end{figure*}

\paragraph{Latent Space Analysis.}
To gain insight into how contextual and parametric information is represented internally, we perform a per-layer 2D principal component analysis (PCA; \citealp{PCA}) on decoder hidden states extracted at the first token of generation (FTG). As shown in \cref{fig:PCA_activations} (refer to \cref{app:more-results} for all the layers), representations from early layers largely overlap, while a clearer separation between contextual and parametric activations emerges from the middle layers onward in all three models. This pattern suggests that attribution-related information becomes increasingly linearly accessible in higher layers, indicating that contextual and parametric signals are likely encoded in separable directions of the hidden state space.

\subsection{Evaluating Bias}\label{sec:bias}
To see if \textsc{AttriWiki} introduces lexical shortcuts between classes as an unintended effect of our knowledge-proxy,\footnote{For instance, generic descriptors such as “a country” may occur more frequently in contexts associated with widely known facts, inadvertently biasing a classifier toward predicting parametric knowledge based on surface-level phrasing rather than attribution.} we train text-only classifiers on the passages. 
A balanced bag-of-words (BoW) logistic regression and a logistic regression model trained on DeBERTaV3\footnote{\texttt{microsoft/deberta-v3-large}.} embeddings \citep{he2021debertav3} are evaluated under five-fold cross-validation (for specifics see \cref{app:hyperparams-bias}). 

\begin{table}[tb]
    \centering
    \small
    \renewcommand*{\arraystretch}{1.2}
    \resizebox{0.825\columnwidth}{!}{
    \begin{tabular}{@{}llr@{}}
    
        \toprule
        \textbf{Model} & \textbf{Classifier} & \textbf{$F_1$} \\
        \midrule
        \multirow{2}{*}{Llama-3.1-8B}
          & BoW        & $.652 \pm .008$\\
          & Embedding  & $.663 \pm .002$\\
        \midrule
        \multirow{2}{*}{Mistral-7B-v0.1}
          & BoW        & $.660 \pm .005$\\
          & Embedding  & $.667 \pm .005$\\
        \midrule
        \multirow{2}{*}{Qwen2.5-7B}
          & BoW        & $.675 \pm .004$\\
          & Embedding  & $.648 \pm .006$\\
        \bottomrule
    \end{tabular}%
    }
\caption{Bias detection performance using BoW and embedding-based classifiers on \textsc{AttriWiki} with  Macro-$F_1$ (mean $\pm$ cross-validation standard deviation). }
\label{tab:bias_detection}
\end{table}

\paragraph{Results.} 
Results are shown in \cref{tab:bias_detection}.
Bag-of-words and embedding-based classifiers achieve $F_1$ scores of 0.65–0.67, indicating the presence of lexical cues, though these alone are insufficient for reliable performance. BoW on the data for all models shows similar performance and nearly identical bigram profiles: contextual examples favor entity-specific terms (e.g., named, born in, towns), whereas parametric examples lean toward broader encyclopedic vocabulary (e.g., countries, nationalities). This suggests that our knowledge proxy introduces a mild and stable lexical bias.

\section{Attribution Classification}
To assess whether hidden states in \textsc{AttriWiki} truly encode the \emph{source} of a model’s knowledge, we train three probing classifiers of increasing complexity. Each model is trained on 64\% of \textsc{AttriWiki}, validated on 16\%, and tested on the remaining 20\%, then evaluated for out-of-domain generalization. Results are reported as macro-$F_{1}$. Consult \cref{app:hyperparams-attri} for more details.

\subsection{Classifiers}
Since the data for Llama and Mistral shows a mild class imbalance ($\approx$ 3{:}2 parametric–contextual), we compensate through loss weights. To prevent information leakage, we further enforce title-disjoint train–test splits, ensuring that no samples originating from the same article appear in both splits.

\paragraph{Final-Layer Logistic Regression (\textit{Final-LR}).}
As a minimal baseline, we train a single logistic unit on the normalized hidden vector $\mathbf{h}_{L} \in \mathbb{R}^{H}$ from the final transformer layer:
\begin{equation}
p = \sigma(\mathbf{w}^{\top}\mathbf{h}_{L} + b),
\end{equation}
where $\mathbf{w} \in \mathbb{R}^{H}$ and $b \in \mathbb{R}$ are the classifier weights and bias, $\sigma(\cdot)$ is the sigmoid function, and training uses an $\ell_2$-regularized logistic loss with inverse-frequency class weighting.

\paragraph{Layer-Weighted Logistic Regression (\textit{Layer-LR}).}
To exploit depth information, we learn a set of unconstrained  parameters $\boldsymbol{\theta}\in\mathbb{R}^{L}$, which are transformed via a softmax to obtain aggregation weights $\boldsymbol{\alpha}=\mathrm{softmax}(\boldsymbol{\theta})$ over transformer layers. These weights are used to form a blended representation
\begin{equation}
\bar{\mathbf{h}}=\sum_{\ell=1}^{L}\alpha_{\ell}\mathbf{h}_{\ell}.
\end{equation}
A logistic head then predicts the probability of parametric retrieval from $\bar{\mathbf{h}}$.  All parameters $(\boldsymbol{\theta},\mathbf{w},b)$ are trained jointly using binary cross-entropy with logits and positive-class reweighting to account for label imbalance, optimized with AdamW \citep{kingma2017adammethodstochasticoptimization}. Hyperparameters and training details are reported in \cref{app:hyperparams-attri}.

\paragraph{Layer-Weighted MLP (\textit{Layer-MLP}).}
We extend the previous model with sparsemax-normalized aggregation weights \citep{martins2016softmaxsparsemaxsparsemodel}, encouraging focus on a small number of salient layers, motivated by the tendency of softmax-aggregated MLP probes to learn diffuse layer weights. The linear classifier is replaced with a compact two-layer MLP operating on the aggregated representation $\bar{\mathbf{h}}$:
\begin{equation}
p=\sigma\!\left(\mathbf{w}_{2}^{\top}\mathrm{GELU}(\mathbf{W}_{1}\bar{\mathbf{h}})+b\right),
\end{equation}
We use GELU \citep{gelu}, where $\mathbf{W}_{1}\in\mathbb{R}^{m\times H}$ projects the hidden representation of dimension $H$ into a bottleneck of size $m$, and $\mathbf{w}_{2}\in\mathbb{R}^{m}$ denotes the output-layer weights. Hyperparameters and training details are reported in \cref{app:hyperparams-attri}.

\subsection{Out-of-distribution Datasets}
\label{section:method-webq-squad}
All classifiers are first trained on \textsc{AttriWiki} and then evaluated on $2{,}000$ samples from two external QA datasets.  The best classifier is additionally compared against the probe of \citet{tighidet2024probinglanguagemodelsknowledge} on conflicting scenarios. All samples include a single demonstration line to enforce output format (see \cref{app:oneshot-prompts}).

\paragraph{\textsc{ParaRel} \citep{elazar-etal-2021-measuring}.}To situate our approach relative to \citet{tighidet2024probinglanguagemodelsknowledge}, we evaluate on prompts derived from \textsc{ParaRel}, a structured dataset of factual \textsc{(subject, relation, object)} triples sourced from Wikidata. Each triple is converted into a probing prompt consisting of a counter-factual statement contradicting the model's parametric knowledge, followed by a query over the same subject--relation pair. Labels are assigned based on whether the model's response matches the planted counter-object (\textsc{CK}) or the true parametric object (\textsc{PK}), other answers are ignored. Relations are grouped into eight semantic categories.

\paragraph{\textsc{WebQuestions} \citep{berant-etal-2013-semantic}.} 
Each factoid question lacks supporting context (e.g., “Who developed general relativity?”~$\rightarrow$~\emph{Albert~Einstein}); hence, answers must come from the model’s parametric memory. Items are wrapped in a one-shot question-answer prompt for compatibility with decoder-only models.

\paragraph{\textsc{SQuAD} \citep{rajpurkar2016squad100000questionsmachine}.}
Passages contain answers verbatim, making the task predominantly \emph{contextual}. To filter memorized facts, we query each question without its paragraph; if the model still answers correctly, the item is recorded as prior knowledge and used for a separate experiment (see below). The remaining examples form the contextual test set.

\paragraph{SQuAD Variants.}
We derive three controlled variants:
(a) an \emph{irrelevant context} split pairing each known question with an irrelevant, randomly sampled paragraph to force parametric retrieval;
(b) an \emph{answer-string decoy} version where the paragraph contains the correct answer verbatim in an unrelated context, testing reliance on surface overlap; and
(c) a \emph{dual-source} version retaining both context and prior-knowledge cases to study retrieval preference.
Together, these variants test robustness against degenerate heuristics, e.g., predicting attribution solely from the presence of context.

\subsection{Results}
All reported macro-$F_1$ values are accompanied by bootstrap standard deviations estimated from $B = 1{,}000$ samples with replacement.

\begin{table}[tb]
    \centering
    \resizebox{\columnwidth}{!}{
    \renewcommand*{\arraystretch}{1.2}
    \begin{tabular}{lrrr}
    \toprule
    \textbf{Model} & \textbf{Final-LR} & \textbf{Layer-LR} & \textbf{Layer-MLP} \\
    \midrule
    \multicolumn{4}{l}{\textit{Token = FTG (First Token of Generation)}} \\
    Llama-3.1-8B& $.841 \pm .005$& $.919 \pm .004$& $.922 \pm .004$\\
    Mistral-7B-v0.1& $.844 \pm .005$& $.922 \pm .004$& $.933 \pm .004$\\
    Qwen2.5-7B& $.835 \pm .005$& $.912 \pm .004$& $.926 \pm .004$\\
    \multicolumn{4}{l}{\textit{Token = LTE (Last Token of Entity)}} \\
    Llama-3.1-8B& $.854 \pm .005$& $.948 \pm .003$& $.955 \pm .003$\\
    Mistral-7B-v0.1& $.904 \pm .004$& $.958 \pm .003$& $.961 \pm .003$\\
    Qwen2.5-7B& $.867 \pm .005$& $.935 \pm .004$& $.949 \pm .006$\\
    \bottomrule
    \end{tabular}
    }
    \caption{Macro-$F_1$ (mean $\pm$ bootstrap std, $B=1000$)  on \textsc{AttriWiki-test} ($n\approx5{,}000$).}
    \label{tab:attriwiki_token_table}
\end{table}

\paragraph{Attribution Performance.}
\cref{tab:attriwiki_token_table} reports attribution classification on the held-out \textsc{AttriWiki} split. We compare \textit{Final-LR}, \textit{Layer-LR}, and \textit{Layer-MLP}. Layer aggregation improves $F_1$ by 7–11 pp over the final-layer baseline, while the MLP yields only marginal gains ($\approx$1 pp). Token choice matters: the last token of the entity (\textit{LTE}) performs best (0.95–0.96), while the first token of generation (\textit{FTG}) remains slightly lower but above 0.90. Mistral consistently outperforms Llama and Qwen by one point, though trends are consistent across models, with Macro-$F_1$ closely matching accuracy.

\begin{table}[tb]
\centering
\renewcommand*{\arraystretch}{1.2}
\resizebox{0.8\columnwidth}{!}{
\begin{tabular}{lrr}
\toprule
\textbf{Model} & \textbf{SQuAD} & \textbf{WebQ} \\
\midrule
Llama-3.1-8B    & $.977 \pm .003$& $.955 \pm .004$\\
Mistral-7B-v0.1& $.918 \pm .005$& $.996 \pm .001$\\
Qwen2.5-7B& $.996 \pm .005$& $.991 \pm .002$\\
\bottomrule
\end{tabular}%
}
\caption{Generalization accuracy (mean $\pm$ bootstrap std, $B=1000$) of Layer-LR (FTG) on out-of-domain QA datasets. WebQ = WebQuestions.}
\label{tab:generalisation_results}
\end{table}

\paragraph{Generalization.}
To assess out-of-distribution generalization, we evaluate the classifier on two out-of-domain QA benchmarks: \textsc{SQuAD} (contextual) and \textsc{WebQuestions} (parametric). As shown in \cref{tab:generalisation_results}, it generalizes near perfectly without fine-tuning, achieving 0.92–0.99 accuracy across Llama, Mistral, and Qwen on both benchmarks. Similar results are observed for the Layer-MLP in the appendix, in \cref{tab:generalisation_results_mlp}. We report only on first-token generation (FTG), because it allows us to analyze incorrect responses in later sections.

\begin{figure}[tb]
    \centering
    \includegraphics[width=\linewidth]{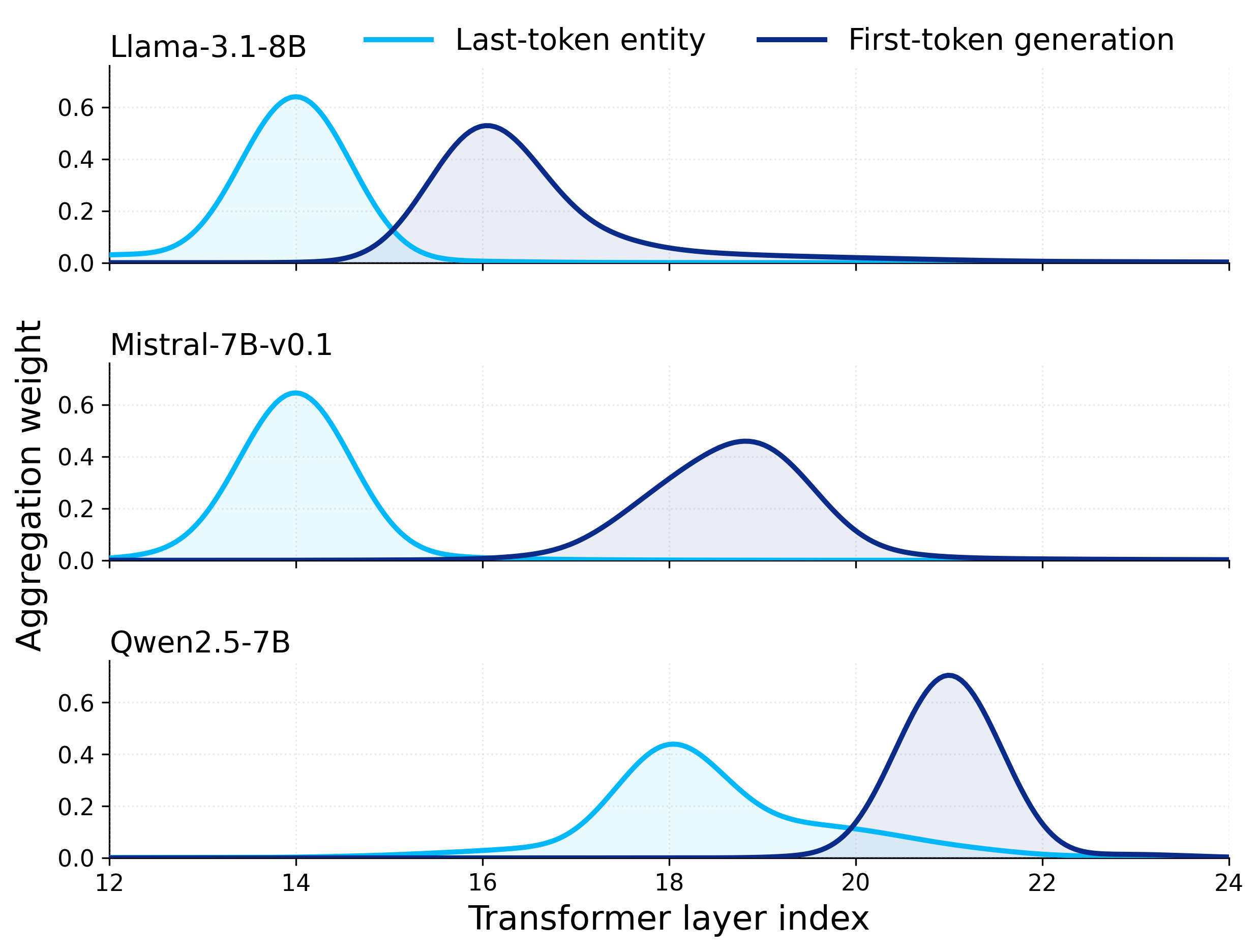}
    \caption{Layer-aggregation weights learned by \textit{Layer-LR} for first-token generation and last-token entity representations. Curves show learned weights across transformer layers, smoothed with a Gaussian kernel for visual clarity. The x-axis is restricted to layers 10--24, as earlier and later layers receive negligible weight.
}
    \label{fig:layer-aggregation-weights}
\end{figure}

\paragraph{Which Layers Does the Classifier Rely on?}
As shown in \cref{fig:layer-aggregation-weights}, the classifier concentrates weight in the lower-middle transformer layers across models when using the last token of the entity. 
For first-token generation, peak contributions occur at layers 16 (Llama-3.1-8B), 19 (Mistral-7B-v0.1), and 21 (Qwen2.5-7B), while last-token entity representations peak earlier at layers 14 (Llama, Mistral) and 18 (Qwen). Notably, Qwen assigns weight to relatively deeper layers given its shallower overall depth (28 versus 32).
This indicates that using the last token of the entity might make the probe focused on entity-specific, lexical information, while the first-token generation hidden states push the probe to focus on potentially high-level information processed in later layers.

\paragraph{Comparison to \citet{tighidet2024probinglanguagemodelsknowledge}.}
\cref{tab:probe_comparison} compares our \textsc{AttriWiki}-trained probe Layer-LR on FTG against the best-performing layer of \citet{tighidet2024probinglanguagemodelsknowledge} evaluated on \textsc{ParaRel} under the same exact scheme. The \textsc{AttriWiki} probe consistently outperforms across all three models on the never-seen before data, with gains ranging from $3$ pp (Qwen2.5-7B) to $13$ pp (Llama-3.1-8B). Unlike \citet{tighidet2024probinglanguagemodelsknowledge}, our probe requires no conflicting context during training, extracts hidden states at the first generated token rather than the last input token, and learns to aggregate across all layers jointly rather than selecting a single layer post-hoc.
We include the per-category results in \cref{app:more-results}.

\begin{table}[tb]
\centering
\renewcommand*{\arraystretch}{1.2}
\resizebox{0.9\columnwidth}{!}{%
\begin{tabular}{llr}
\toprule
\textbf{Model} & \textbf{Probe} & Macro-$F_1$ \\
\midrule
\multirow{2}{*}{Mistral-7B-v0.1} & \citeauthor{tighidet2024probinglanguagemodelsknowledge} (L10) & $.745 \pm .014$ \\
                                  & \textsc{AttriWiki} &\textbf{ $.806 \pm .012$} \\
\midrule
\multirow{2}{*}{Llama-3.1-8B}    & \citeauthor{tighidet2024probinglanguagemodelsknowledge} (L11) & $.640 \pm .019$ \\
                                  & \textsc{AttriWiki} & \textbf{$.771 \pm .016$} \\
\midrule
\multirow{2}{*}{Qwen2.5-7B}      & \citeauthor{tighidet2024probinglanguagemodelsknowledge} (L22) & $.757 \pm .007$ \\
                                  & \textsc{AttriWiki} & \textbf{$.793 \pm .006$} \\
\bottomrule
\end{tabular}%
}
\caption{Comparison of \citeauthor{tighidet2024probinglanguagemodelsknowledge} and \textsc{AttriWiki} probe (Layer-LR on FTG) across models with Macro-$F_1$ (mean $\pm$ bootstrap std, $B=1000$). Layer number in parentheses indicates the best-performing layer.}
\label{tab:probe_comparison}
\end{table}

\subsection{Ablation}


Ablations are performed on Layer-LR at FTG, as it is robust and does not rely on entity mentions.

\paragraph{Degenerate Attribution.}
We evaluate whether attribution probes rely on degenerate heuristics rather than genuine provenance signals. 
Firstly, we test the answer-string decoy variant of SQuAD, where the gold answer appears verbatim in an unrelated context—performance drops for all classifiers. 
Nevertheless, \textit{Layer-LR} remains well above chance (0.724)\footnote{Semantic irrelevance is not explicitly verified; we use this ablation to uncover potential model shortcuts.}. In contrast, the MLP collapses to near-random performance (0.541), indicating that the MLP overfits to lexical shortcuts such as entity repetition, whereas \textit{Layer-LR} captures a more robust attribution signal (see \cref{tab:decoy} in \cref{app:more-results}). 
Secondly, in the irrelevant-context setting---where known facts are paired with unrelated passages---92\% of answers are still attributed as parametric, ruling out a trivial heuristic that equates the mere presence of context with contextual attribution.
Thirdly, when both parametric memory and context suffice to answer a question, models overwhelmingly default to the contextual channel (87.1\%). This rules out the explanation that the probe merely detects whether a fact is stored in the model’s parameters, rather than the knowledge source. 

\paragraph{Error and Attribution Mismatch.}
To analyze the relationship between attribution and answer correctness, we consider two complementary conditions: (i) present parametric knowledge paired with irrelevant context, and (ii) absent parametric knowledge paired with relevant context.
We use \textsc{SQuAD} with random contexts in case of parametric knowledge and original context otherwise, recording answer correctness.
Source alignment (parametric when required, or contextual when required) and correctness form a $2{\times}2$ contingency table, analyzed using Fisher’s exact test \citep{10.1111/j.2397-2335.1935.tb04208.x}, with relative risk as the effect size.
Attribution mismatches strongly correlate with errors ($p \ll 0.001$), with asymmetric effects: relying on misleading context increases errors by up to $\approx$70\% when parametric knowledge is required, whereas defaulting to parametric memory in contextual settings increases errors by only $\approx$30\%.

\section{Discussion}
Our findings indicate that \emph{contributive attribution}, whether a completion is driven by contextual evidence or parametric memory, is a representation-level property that is accessible from hidden states, offering a diagnostic lens on knowledge channel usage that complements hallucination taxonomies distinguishing faithfulness from factuality errors \citep{survey_taxonomy_hallucinations, qi2024catchelephantsurveyevolvement, ye2023cognitivemiragereviewhallucinations}. Our mismatch analysis motivates this: using the wrong knowledge source substantially increases error risk, particularly when misleading context overrides parametric knowledge. Yet many errors persist despite correct source alignment, indicating that attribution is one diagnostic axis among others rather than a complete error detector.

Moreover, attribution is learnable even without explicit knowledge conflicts, as evidenced by our strong out-of-distribution performance including on unseen conflict settings. This challenges conflict-based approaches \citep{tighidet2024probinglanguagemodelsknowledge, xu2024knowledgeconflictsllmssurvey}, which conflate attribution with conflict resolution by design. Where prior mechanistic work shows that contextual and parametric signals coexist as superposed activations routed through largely distinct circuits \citep{zhao2025understanding}, our results go further: their relative contribution is quantitatively detectable at generation time without ever exposing contradictory information, suggesting conflict-free training is not merely sufficient but actually preferable.

Two further results confirm prior work. The limited gains from an MLP head and its degradation under the answer-string decoy suggest that added model complexity encourages shortcut learning, while linear probes more reliably predict provenance---consistent with the knowledge source being another property encoded linearly in hidden representations \citep{park2023linear, jiang2024origins, merullo2025linear}. 
The preference of the probe for upper-middle layers aligns with prior findings that mid-layer activations are most informative for knowledge attribution \citep{tighidet2024probinglanguagemodelsknowledge}, and with broader evidence that later layers encode decision-relevant abstractions beyond surface form \citep{jawahar-etal-2019-bert, tenney2019learncontextprobingsentence}.

\section{Conclusions}
We study contributive attribution in large language models using a self-supervised, conflict-free pipeline (\textsc{AttriWiki}). Linear probes trained on \textsc{AttriWiki} achieve up to 0.96 $F_1$ on Llama, Mistral, and Qwen models and generalize strongly to SQuAD, WebQuestions and unseen conflict settings. Attribution mismatches increase error risk by 30–70\%, particularly under misleading contexts, although correct attribution alone does not guarantee correctness. Together, these findings establish attribution as a meaningful diagnostic signal.

Future work should explore how attribution signals can be integrated into downstream systems. In retrieval-augmented generation (RAG; \citealp{RAGpretrain, RAGTrillionTokens}), attribution probes could provide online signals to detect ignored evidence and trigger selective verification or re-retrieval. 
In LLM-powered chatbots, exposing attribution may also elicit more critical user engagement by clarifying whether responses rely on retrieved evidence or prior knowledge, similar to \citet{deng2023prompting}. Such applications call for human evaluation, both to validate attribution judgments independently of our automated proxy and to establish whether users find the signal actionable in practice. 
Other works have identified potential interventions through steering model activations along a knowledge-source direction \citep{basu2025mechanistic, bi2025parameters}.


\section*{Limitations}
\paragraph{Data.}
 \textsc{AttriWiki}’s controlled Wikipedia domain may limit generalization; testing attribution signals on legal text, dialogue, and noisy web data is essential. Extension to multilingual settings also poses challenges to \textsc{AttriWiki} since knowledge varies across languages \citep{kassner-etal-2021-multilingual}. Additionally, shortcut learning poses a risk, inputs might be lexically associated with parametric knowledge as demonstrated in \cref{sec:bias}. In future work, such shortcuts could be mitigated at the dataset level through adversarial filtering, which iteratively removes examples solved by a lexical-only classifier, such as our BoW baseline \citep{Debiasdataset}.
 Furthermore, distribution shift remains a concern: \textsc{ParaRel} covers Wikidata subjects not present in\textsc{ AttriWiki}, and per-category results in \cref{app:more-results} show that probe performance varies greatly across categories, suggesting sensitivity to unseen domains.
 
 \paragraph{Attribution Classification.}
The classifier would benefit from moving beyond token-level attribution to sequence-level approaches. While our current method performs well on factual questions, it struggles in more natural settings where no clear entity span exists. Aggregation strategies, such as attention-based span scoring, could provide more accurate attribution in these cases.
Another limitation is model dependence: each LLM requires a new classifier, and retraining the model often necessitates retraining the classifier and regenerating data if the model’s knowledge evolves.

\paragraph{Ablation.}
Although informative, error analysis remains challenging. Phenomena such as contradictions, exaggerations, and refusal to answer are often grouped under hallucinations, yet our focus here is on forced factual retrieval. Assessing answer correctness is equally difficult. Minor variations, such as one-letter acronym slips (e.g., GDP $\rightarrow$ GPD), can bypass the cross-encoder (\emph{synonym matching}), while valid polarity with low lexical overlap may also lead to misclassifications. For example, the answer ``smaller'' correctly responds to ``Were the houses bigger or smaller?'' but fails against the gold label “smaller houses.” 

\section*{Ethics Considerations}
This work only uses publicly available datasets and does not involve human subjects or personal data. One API-based LLM is used solely for data generation. We do not identify significant ethical risks beyond those common to interpretability research.

\section*{Acknowledgments}
This work is supported by the Dutch National Science Foundation (NWO Vici VI.C.212.053) and KPMG NL. We would also like to express our gratitude to Ivan Titov for his valuable guidance and feedback throughout this work. 

\bibliography{custom}
\newpage
\appendix

\section{Resources}\label{app:resources}

This appendix reports the computational cost of the work presented here, including
exploratory and discarded runs. We distinguish three quantities: the total
cost of the research program, the cost of reproducing the pipeline, and the
marginal cost of extending it to an additional target model. 

\subsection{Data Generation (API)}

Total \textsc{gpt-4o-mini} usage across the project was approximately 165M input tokens and 17M output tokens ($\approx \$35$).
This figure covers the complete research process: exploratory runs, prompt
iteration, discarded configurations, and an initial sample of Wikipedia pages that
was later superseded. It is substantially larger than the cost of the final
pipeline and should be read as the cost of \emph{arriving at} \textsc{AttriWiki},
not of building it.
Two narrower figures are more informative for practitioners.

\paragraph{Extending to a New Target Model.} Because entity selection and
knowledge-test prompt generation are model-independent, adding a target model to
an existing entity set requires only the model-specific stages, namely entity
removal and completion-cue generation, which are conditioned on which entities the
model recalls. We measured this directly when adding \textsc{Qwen2.5-7B}: 7.9M
input and 1.9M output tokens, or \$2.33 at the pricing in effect at the time. We
additionally reused removals across models wherever their known/unknown labels
coincided, so a model whose parametric coverage overlaps less with the existing
set would incur a somewhat higher cost.

\paragraph{Building From a New Corpus.} A clean build on a new data source must
re-run the model-independent stages. We estimate this at roughly 80M input tokens and 10M output tokens for a corpus comparable to our 20k Wikipedia passages. The estimate is
dominated by knowledge-test generation, which passes a
full passage as input for every passage and every candidate entity and a knowledge question as output; the
model-specific stages contribute under 10M tokens by the measurement above.

\subsection{Target Model Inference and Probe Training (GPU)}

Knowledge testing, hidden-state extraction, and probe training were run on a single
NVIDIA A100. Knowledge testing dominates GPU cost, as it evaluates
three prompt formats for every candidate entity across the sampled passages, taking
up to 7 hours per target model on a single GPU. Hidden-state extraction over the rewritten passages
is cheaper, requiring up to 3 hours per model despite capturing full-depth hidden
states at two token positions, as it involves a single pass over the constructed
prompts. Total inference cost is therefore bounded at roughly 10 GPU-hours per
target model, or 30 GPU-hours across the three models. Probe training
is negligible by comparison: a single \textit{Layer-LR} probe converges in under
a minute.

Following
\citet{co2calculator}, 30 GPU-hours on the A100 draws approximately 12\,kWh,
which at a carbon intensity of 0.268\,kg/kWh for the Dutch grid and a data center
PUE of 1.19 corresponds to roughly 3.8\,kg CO\textsubscript{2}eq.
\subsection{Cost Profile of the Method}

The cost structure of attribution probing is favorable, and we consider this a
substantive property of the approach rather than an incidental one. The expensive
stages are incurred once: dataset construction and hidden-state extraction are
one-off, and the model-independent portion is amortized across every target model
subsequently probed. What remains at deployment is a linear probe over hidden
states that are already computed during the forward pass, adding negligible
inference overhead. This contrasts with sampling-based hallucination detection
such as SelfCheckGPT \citep{manakul2023selfcheckgptzeroresourceblackboxhallucination}, which requires multiple full
generations per query, and with source-aware training \citep{khalifa2024sourceawaretrainingenablesknowledge},
which requires retraining the model itself.

The principal recurring cost is model dependence, as noted in the Limitations:
each target model requires its own probe, and a model update requires regenerating
the model-specific data and retraining. The measurement above bounds this at a few
dollars of API usage plus several GPU-hours per model, which we regard as
acceptable for a diagnostic that is otherwise free at inference time.

\section{Hyperparameters}\label{app:hyperparams-bias}

This appendix documents all hyperparameters used in our experiments.
We separate configurations for \emph{bias evaluation} models and
\emph{attribution classifiers}.
\subsection{Bias evaluation}
We evaluate bias using two complementary classifiers: a sparse
BoW model and an embedding-based model.
The former serves as a lightweight lexical baseline, while the latter
tests whether bias signals persist in contextualized representations.
The corresponding hyperparameters given in \cref{tab:bias-lr,tab:bias-emb}, respectively.

\begin{table}[h]
\centering
\footnotesize
\renewcommand*{\arraystretch}{1.4}
\setlength{\tabcolsep}{3pt}
\begin{tabular}{p{0.36\columnwidth} p{0.56\columnwidth}}
\hline
\textbf{Component} & \textbf{Configuration} \\
\hline
Text representation
& TF--IDF \newline
  ngram\_range = (1,2) \newline
  max\_features = 5000 \\
Classifier
& Logistic regression \newline
  class\_weight = balanced \newline
  max\_iter = 1000 \newline
  random\_state = 42 \\
\hline
\end{tabular}
\caption{Hyperparameters for the \textbf{TF--IDF logistic regression} bias classifier.}
\label{tab:bias-lr}
\end{table}

\begin{table}[h]
\renewcommand*{\arraystretch}{1.4}
\centering
\footnotesize
\setlength{\tabcolsep}{3pt}
\begin{tabular}{p{0.36\columnwidth} p{0.56\columnwidth}}
\hline
\textbf{Component} & \textbf{Configuration} \\
\hline
Text representation
& Transformer embeddings \newline
  model = microsoft/deberta-v3-large \newline
  max\_length = 128 \newline
  padding = max\_length \newline
  truncation = true \newline
  repr. = [CLS] \newline
  dtype = float16 \newline
  eval mode, no gradients \\
Classifier
& Logistic regression \newline
  max\_iter = 1000 \newline
  random\_state = 42 \\
\hline
\end{tabular}
\caption{Hyperparameters for the \textbf{embedding-based} bias classifier.}
\label{tab:bias-emb}
\end{table}

\subsection{Attribution classifiers}
\label{app:hyperparams-attri}
Attribution classifiers are trained to probe internal model
representations. Hyperparameters for \textbf{Layer-LR} and \textbf{Layer-MLP} were selected via grid search over dropout $p$
($\{0,0.1,0.2\}$), weight decay $\lambda$
($\{0,5\!\times\!10^{-4},10^{-3},2\!\times\!10^{-3}\}$), learning rate $\eta$
($\{5\!\times\!10^{-4},10^{-3},2\!\times\!10^{-3}\}$), and (for Layer-MLP) bottleneck size $m$
($\{64,128\}$), using a title-aware 15\% validation split.
For the grid search, models were trained for 10 epochs; the selected configuration was then trained for up to 100 epochs with early stopping on validation macro-$F_1$
(tie-breaker: validation accuracy). 
Selected values are reported in \cref{tab:final_lr,tab:layer_lr,tab:layer_mlp}.

\begin{table}[tb]
\centering
\footnotesize
\setlength{\tabcolsep}{3pt}
\renewcommand*{\arraystretch}{1.4}
\begin{tabular}{p{0.36\columnwidth} p{0.56\columnwidth}}
\hline
\textbf{Component} & \textbf{Configuration} \\
\hline
Input representation
& Last transformer layer ($X[:, -1, :]$) \\
Preprocessing
& StandardScaler (default) \\
Classifier
& Logistic regression \newline
  class\_weight = balanced \newline
  solver = lbfgs \newline
  penalty = $\ell_2$ \newline
  max\_iter = 1000 \\
\hline
\end{tabular}
\caption{Hyperparameters for \textbf{Final-LR} (last-layer logistic regression).}
\label{tab:final_lr}
\end{table}

\begin{table}[tb]
\centering
\footnotesize
\renewcommand*{\arraystretch}{1.4}
\setlength{\tabcolsep}{3pt}
\begin{tabular}{p{0.36\columnwidth} p{0.56\columnwidth}}
\hline
\textbf{Component} & \textbf{Configuration} \\
\hline
Input representation
& Layer stack $X \in \mathbb{R}^{N \times L \times H}$ \newline
  weighted aggregation \\
Layer weighting
& Softmax($\boldsymbol{\theta}$), $\sum_\ell \alpha_\ell = 1$ \\
Normalization
& $\ell_2$ per layer \\
Dropout
& $p = 0.1$ \\
Classifier head
& Linear($H \rightarrow 1$) \\
Loss
& BCEWithLogitsLoss \newline
  pos\_weight $= \#\text{neg} / \#\text{pos}$ \\
Optimizer
& AdamW \newline
  learning rate $\eta = 2\!\times\!10^{-3}$ \newline
  weight decay $\lambda = 10^{-3}$ \\
Training
& batch size $= 64$ \newline
  max epochs $= 100$ \newline
  early stopping (macro-$F_1$, patience $= 3$) \newline
  title-aware validation split ($0.15$, seed $= 42$) \\
\hline
\end{tabular}
\caption{Hyperparameters for \textbf{Layer-LR} (layer-weighted linear classifier).}
\label{tab:layer_lr}
\end{table}

\begin{table}[tb]
\centering
\footnotesize
\renewcommand*{\arraystretch}{1.4}
\setlength{\tabcolsep}{3pt}
\begin{tabular}{p{0.36\columnwidth} p{0.56\columnwidth}}
\hline
\textbf{Component} & \textbf{Configuration} \\
\hline
Input representation
& Layer stack $X \in \mathbb{R}^{N \times L \times H}$ \newline
  weighted aggregation \\
Layer weighting
& Sparsemax($\boldsymbol{\theta}$) \\
Normalization
& $\ell_2$ per layer \\
MLP head
& Linear($H \rightarrow m$), $m = 64$ \newline
  GELU, Dropout($p = 0.1$) \newline
  Linear($m \rightarrow 1$) \\
Loss
& BCEWithLogitsLoss \newline
  pos\_weight $= \#\text{neg} / \#\text{pos}$ \\
Optimizer
& AdamW \newline
  learning rate $\eta = 10^{-3}$ \newline
  weight decay $\lambda = 10^{-3}$ \\
Training
& batch size $= 64$ \newline
  max epochs $= 100$ \newline
  early stopping (macro-$F_1$, patience $= 3$) \newline
  title-aware validation split ($0.15$, seed $= 42$) \\
\hline
\end{tabular}
\caption{Hyperparameters for \textbf{Layer-MLP} (layer-weighted MLP classifier).}
\label{tab:layer_mlp}
\end{table}

\section{Additional Results}
\label{app:more-results}

\subsection{Latent Space Analysis}\label{app:latent-space}

\begin{figure*}[tb]
    \centering
    \includegraphics[width=0.75\textwidth]{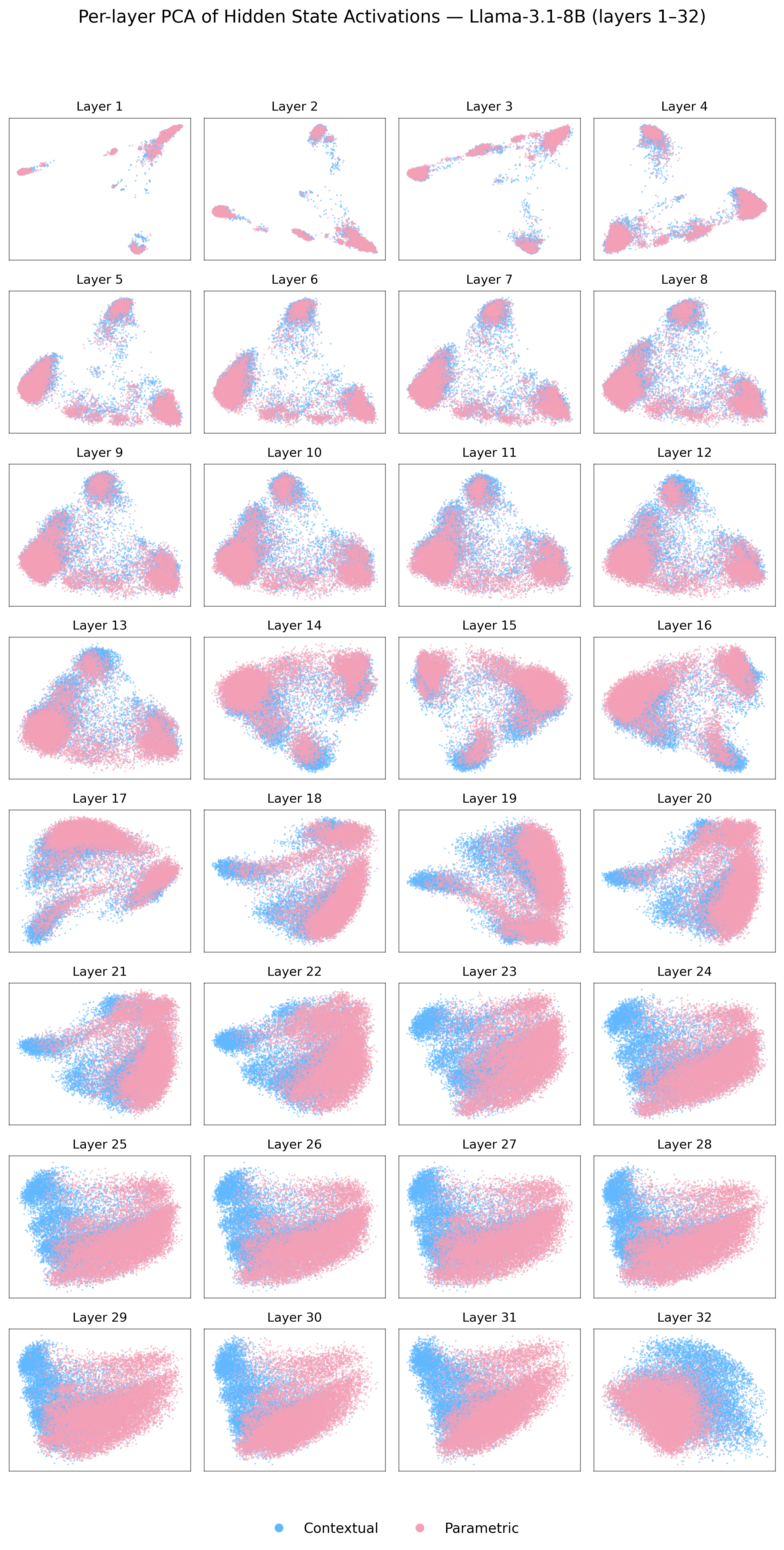}
    \caption{Per-layer PCA of hidden states for Llama-3.1-8B.}
    \label{fig:layer_pca_llama}
\end{figure*}

\begin{figure*}[tb]
    \centering
    \includegraphics[width=0.75\textwidth]{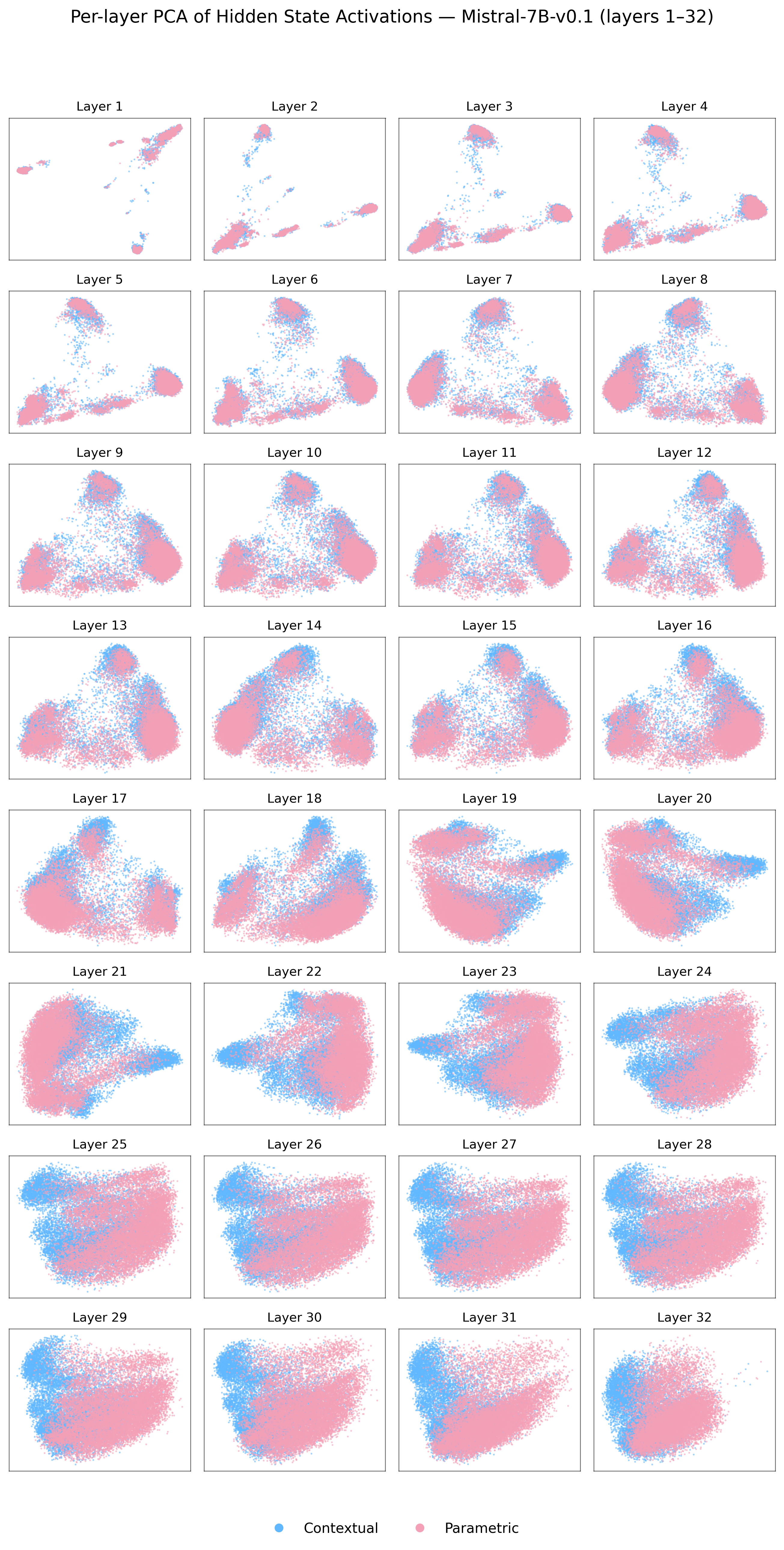}
    \caption{Per-layer PCA of hidden states for Mistral-7B-v0.1.}
    \label{fig:layer_pca_mistral}
\end{figure*}

\begin{figure*}[tb]
    \centering
    \includegraphics[width=0.75\textwidth]{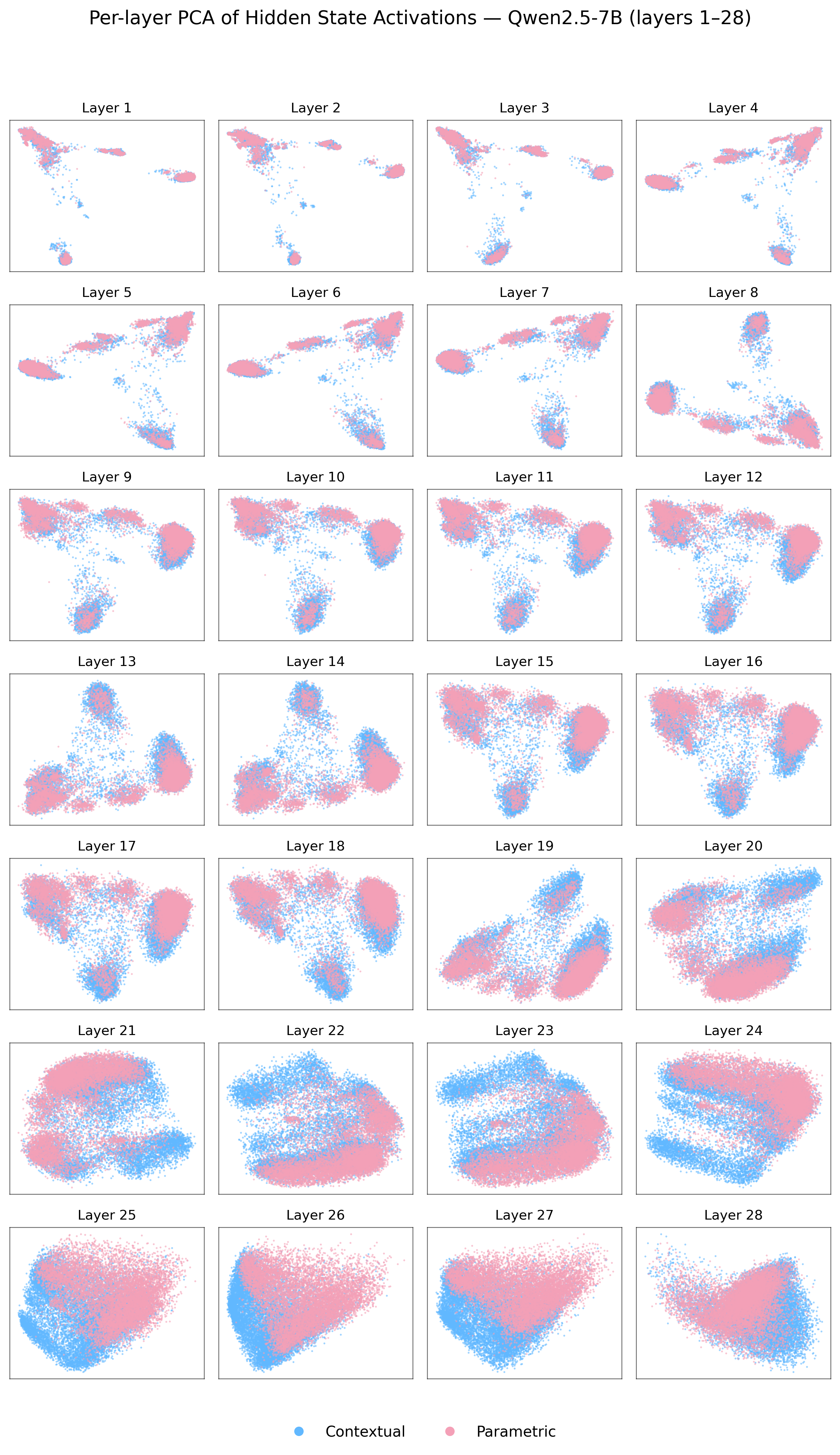}
    \caption{Per-layer PCA of hidden states for Qwen2.5-7B.}
    \label{fig:layer_pca_qwen}
\end{figure*}

Here we report the full latent space plots from \cref{fig:PCA_activations} for all models, with \cref{fig:layer_pca_llama} for Llama-3.1-8B, \cref{fig:layer_pca_mistral} for Mistral-7B-v0.1, and \cref{fig:layer_pca_qwen} for Qwen2.5-7B.

\subsection{Out-of-distribution and Generalization Results}

This section reports additional experimental results that support the
main findings in the paper, including ablation studies and
out-of-domain generalization performance. 
\cref{tab:decoy} shows the accuracy for the answer-string decoy variant for \textsc{SQuAD}, where we can see that the MLP variant of the probe almost collapses to random accuracy, suggesting potential overfitting.
\cref{tab:generalisation_results_mlp} shows the results for the MLP probe on \textsc{SQuAD} and \textsc{WebQuestions}.

\begin{table}[htb]
\centering
\renewcommand*{\arraystretch}{1.2}
\resizebox{0.55\columnwidth}{!}{
\begin{tabular}{lc}
\toprule
\textbf{Classifier} & \textbf{Accuracy} \\
\midrule
Layer-LR & .724 \\
Layer-MLP   & .541 \\
\bottomrule
\end{tabular}
}
\caption{Answer-string decoy ablation on \textsc{SQuAD} using Llama-3.1-8B.
While performance drops for all classifiers, the linear probe remains
well above chance, whereas the MLP collapses to near-random accuracy,
indicating reliance on entity repetition.}
\label{tab:decoy}
\end{table}

\begin{table}[htb]
\centering
\renewcommand*{\arraystretch}{1.2}
\resizebox{0.8\columnwidth}{!}{
\begin{tabular}{lrr}
\toprule
\textbf{Model} & \textbf{SQuAD} & \textbf{WebQ} \\
\midrule
Llama-3.1-8B     & .972 & .959 \\
Mistral-7B-v0.1  & .958 & .948 \\
Qwen2.5-7B& .998 & .980 \\
\bottomrule
\end{tabular}
}
\caption{Generalization accuracy of Layer-MLP (FTG) on out-of-domain QA
datasets. WebQ = WebQuestions.}
\label{tab:generalisation_results_mlp}
\end{table}

\paragraph{Comparison to \citet{tighidet2024probinglanguagemodelsknowledge}.} In the following \cref{tab:per-group-mistral-7b-v01,tab:per-group-llama-31-8b,tab:per-group-qwen25-7b} we compare the per-group macro-$F_1$ of \citet{tighidet2024probinglanguagemodelsknowledge}'s best-layer probe against the \textsc{AttriWiki} Layer-LR probe on FTG across all three models. 
Relation groups well-represented in encyclopedic Wikipedia text, such as Geographic and Media, transfer well, with \textsc{AttriWiki} consistently matching or exceeding \citeauthor{tighidet2024probinglanguagemodelsknowledge}'s performance. The Corporate group is a consistent outlier. This reflects the fact that \textsc{AttriWiki}'s training data contains essentially no product-ownership facts. These results confirm that domain shift does affect the \textsc{AttriWiki} probe, but only in a targeted and predictable way: performance degrades specifically for themes absent from Wikipedia-style factual text.

\begin{table}[htb]
\centering
\small
\renewcommand*{\arraystretch}{1.2}
\resizebox{0.98\columnwidth}{!}{
\begin{tabular}{lrrr}
 \toprule
 \textbf{Relation group} & $n$ & \textbf{\citeauthor{tighidet2024probinglanguagemodelsknowledge} (L10)} & \textsc{AttriWiki} \\
 \midrule
 Corporate & 100 & .333 $\pm$ .023 & .369 $\pm$ .035 \\
 Geographic & 920 & .780 $\pm$ .014 & .837 $\pm$ .012 \\
 Media & 16 & .329 $\pm$ .057 & .800 $\pm$ .104 \\
 \bottomrule
\end{tabular}%
}
\caption{Per-group macro-$F_1$ (mean $\pm$ bootstrap std, $B=1000$) for \textbf{Mistral 7B}. \citeauthor{tighidet2024probinglanguagemodelsknowledge} uses the best overall layer (L10).}
\label{tab:per-group-mistral-7b-v01}
\end{table}

\begin{table}[htb]
\centering
\small
\resizebox{0.98\columnwidth}{!}{
\begin{tabular}{lrrr}
 \toprule
 \textbf{Relation group} & $n$ & \textbf{\citeauthor{tighidet2024probinglanguagemodelsknowledge} (L11)} & \textsc{AttriWiki} \\
 \midrule
 Corporate & 52 & .411 $\pm$ .061 & .541 $\pm$ .070 \\
 Geographic & 530 & .670 $\pm$ .021 & .763 $\pm$ .019 \\
 Media & 50 & .566 $\pm$ .075 & .878 $\pm$ .046 \\
 Occupy-position & 16 & .328 $\pm$ .058 & 1.000 $\pm$ .000 \\
 Play-instrument & 20 & .589 $\pm$ .113 & .897 $\pm$ .070 \\
 \bottomrule
\end{tabular}%
}
\caption{Per-group macro-$F_1$ (mean $\pm$ bootstrap std, $B=1000$) for \textbf{Llama 3.1 8B}. \citeauthor{tighidet2024probinglanguagemodelsknowledge} uses the best overall layer (L11).}
\label{tab:per-group-llama-31-8b}
\end{table}

\begin{table}[htb]
\centering
\small
\resizebox{0.98\columnwidth}{!}{
\begin{tabular}{lrrr}
 \toprule
 \textbf{Relation group} & $n$ & \textbf{\citeauthor{tighidet2024probinglanguagemodelsknowledge} (L22)} & \textsc{AttriWiki} \\
 \midrule
 Corporate & 640 & .762 $\pm$ .017 & .633 $\pm$ .019 \\
 Geographic & 2928 & .754 $\pm$ .008 & .812 $\pm$ .007 \\
 Hierarchy & 46 & .686 $\pm$ .073 & .768 $\pm$ .062 \\
 Media & 178 & .798 $\pm$ .029 & .886 $\pm$ .024 \\
 Occupy-position & 136 & .624 $\pm$ .042 & .837 $\pm$ .031 \\
 Play-instrument & 20 & .329 $\pm$ .053 & .303 $\pm$ .054 \\
 Religion & 208 & .794 $\pm$ .029 & .840 $\pm$ .025 \\
 \bottomrule
\end{tabular}%
}
\caption{Per-group macro-$F_1$ (mean $\pm$ bootstrap std, $B=1000$) for \textbf{Qwen2.5 7B}. \citeauthor{tighidet2024probinglanguagemodelsknowledge} uses the best overall layer (L22).}
\label{tab:per-group-qwen25-7b}
\end{table}

\onecolumn
\section{Prompts}
\subsection{\textsc{AttriWiki}}
\label{app:prompts}

\subsubsection{Entity selection.}
\label{app:entity_selection}
This prompt was designed to select up to three entities to prevent longer passages containing more entities from being overrepresented in the data.
\begin{examplebox}{Prompt: Entity Selection}
{\small
Given a Wikipedia passage and a list of mentioned entities, you are tasked to narrow down the list of entities to a maximum of three using the following criteria.
\begin{enumerate}
  \item Remove entities that refer to a number, e.g., ``first'', ``last'', ``twelve'', ``fourth''.
  \item Remove entities that refer to a very specific date, e.g., a range, months, days etc. Just a year is acceptable.
  \item Remove entities that refer to any quantity.
  \item The resulting entities should contain both well-known examples and lesser-known examples.
\end{enumerate}
The string of the remaining entities must be exactly the same.\\
\textbf{The passage:}

\texttt{\{text\}}\vspace{0.5em}\\
\textbf{The entities:}

\texttt{\{entities\}}
\vspace{0.5em}\\
\textbf{Maximum of three selected entities:}
}
\end{examplebox}
\newpage
\subsubsection{Knowledge testing.}\label{app:knowledge}
Given a Wikipedia passage and an entity, we present examples of the three knowledge tests.
\begin{enumerate}
    \item Alice: I can't remember exactly who was the king of England in 1265 during the Battle of Evesham. I can't remember. \\
Bob: Actually, I know. It was \textbf{\{King Henry III.\}}
    \item Q: Who was the king of England during the Battle of Evesham in the 13th century? \\
A: \textbf{\{King Henry III\}}
    \item The Battle of Evesham ( 4 August 1265 ) was one of the two main battles of 13th century England 's Second Barons ' War . It marked the defeat of Simon de Montfort, Earl of Leicester , and the  rebellious barons by Prince Edward - later King Edward I - who led the forces of his father , \textbf{\{King Henry III.\}}
\end{enumerate}

\textbf{1. Dialogue}
\begin{examplebox}{Prompt: Knowledge Test (Alice-Bob Dialogue)}
{\small

You will receive a Wikipedia passage of an arbitrary topic and an entity that is mentioned somewhere within the passage. You will create a dialogue between Alice and Bob. \textbf{Alice can't think of the name of} \texttt{[entity]}. She describes it perfectly using the Wikipedia passage. \textbf{Bob is all-knowing}, and tells Alice the name of the entity.

\begin{itemize}
  \item Alice is \textbf{not allowed} to say the name of the entity or part of the entity.
  \item Alice \textbf{must only use information} provided in the Wikipedia passage to describe said entity.
  \item Bob \textbf{must say the exact name} of the entity.
\end{itemize}

Here is an example.

\textbf{Wikipedia passage:}
\begin{quote}
\texttt{< The Battle of Evesham ( 4 August 1265 ) was one of the two main battles of 13th century England 's Second Barons ' War . It marked the defeat of Simon de Montfort , Earl of Leicester , and the  rebellious barons by Prince Edward - later King Edward I - who led the forces of his father , King Henry III . It took place on 4  August 1265 , near the town of Evesham , Worcestershire . >}
\end{quote}

\textbf{Entity:}
\begin{quote}
\texttt{< King Henry III >}
\end{quote}

\textbf{Alice-Bob conversation:}
\begin{quote}
\texttt{< Alice: I can't remember exactly who was the king of England in 1265 during the Battle of Evesham. I can't remember. \\
Bob: Actually, I know. It was King Henry III. >}
\end{quote}

Now it is your turn:

\textbf{Wikipedia passage:}
\begin{quote}
\texttt{< \{passage\} >}
\end{quote}

\textbf{Entity:}
\begin{quote}
\texttt{< \{entity\} >}
\end{quote}

\textbf{Alice-Bob conversation (ending with \texttt{>}):}
\begin{quote}
\texttt{<}
\end{quote}
}
\end{examplebox}
\newpage
\textbf{2. QA-style} \\
\begin{examplebox}{Prompt: Knowledge Test (Question-Answer Style)}
{\small
You will receive a Wikipedia passage of an arbitrary topic and an entity that is mentioned somewhere within the text. Like so:

\textbf{Wikipedia passage:}
\begin{quote}
\texttt{< The Battle of Evesham ( 4 August 1265 ) was one of the two main battles of 13th century England 's Second Barons ' War. It marked the defeat of Simon de Montfort, Earl of Leicester, and the \\rebellious barons by Prince Edward - later King Edward I - who led the forces of his father, King Henry III. It took place on 4 August 1265, near the town of Evesham, Worcestershire. >}
\end{quote}

\textbf{Entity:}
\begin{quote}
\texttt{< King Henry III >}
\end{quote}

\textbf{Question-Answer pair:}
\begin{quote}
\texttt{< Q: Who was the king of England during the Battle of Evesham in the 13th century? \\
A: King Henry III >}
\end{quote}
Follow the same pattern using a new Wikipedia passage. You generate a question (Q) to which the answer is the entity (A).

\textbf{Now it is your turn:}

\textbf{Wikipedia passage:}
\begin{quote}
\texttt{< \{passage\} >}
\end{quote}

\textbf{Entity:}
\begin{quote}
\texttt{< \{entity\} >}
\end{quote}

\textbf{Question-Answer (ending with \texttt{>}):}
\begin{quote}
\texttt{<}
\end{quote}
}
\end{examplebox}
Note that the trailing entity mentions are removed using an exact string match in a later phase. \\

\textbf{3. Truncated passage.} The truncated passage is an exact string match, removing the first mention of the entity in the Wikipedia passage and removing all text succeeding and including the entity.
\newpage
\subsubsection{Entity removal.}
\label{app:entity_removal}
\begin{examplebox}{Prompt: Entity Removal}
{\small
Remove every explicit mention of the entity and any variant (abbreviation, nickname, unambiguous pronoun) from the passage while preserving factual content and fluency.\\
\textbf{One illustrative example:}

\textbf{Entity:}
\begin{quote}
\texttt{United Kingdom}
\end{quote}

\textbf{Original passage:}
\begin{quote}
Winston Churchill was a British statesman, soldier, and writer who served as Prime Minister of the United Kingdom from 1940 to 1945 and again from 1951 to 1955. He led Britain to victory in the Second World War. Among the British public, he is widely considered the greatest Briton of all time. He was born to an aristocratic family in Oxfordshire, England.
\end{quote}

\textbf{Rewritten passage:}
\begin{quote}
Winston Churchill was a statesman, soldier, and writer who served as Prime Minister from 1940 to 1945 and again from 1951 to 1955. He led the nation to victory in the Second World War. Among the public, he is widely considered one of the greatest leaders of all time. He was born to an aristocratic family in Oxfordshire.
\end{quote}

\textbf{Constraints:}
\begin{itemize}
    \item Do \textbf{not} insert placeholder tokens such as \texttt{"\_\_\_\_\_"} or \texttt{"[ENTITY]"}.
    \item \textbf{Forbidden filler words:} \texttt{another, other, elsewhere, someplace, \\something, someone, notable, major, region, and global}.
    \item If deleting the entity breaks a sentence, repair the grammar so the passage would pass professional copy-edit.
    \item Keep all dates, numbers, and named entities that do \textbf{not} refer to the target entity.
    \item If the entity acts as an adjective (e.g., ``UK policy''), rewrite only that phrase so the head noun remains (e.g., ``the policy'').
    \item \textbf{Meta-instruction:} Do \textbf{not} mimic specific wording from the example; use whatever phrasing fits the original passage.
\end{itemize}

\vspace{1em}

\textbf{Now your turn}

\textbf{Entity:}
\begin{quote}
\texttt{\{entity\}}
\end{quote}

\textbf{Original passage:}
\begin{quote}
\texttt{< \{passage\} >}
\end{quote}

\textbf{Rewritten passage (end with \texttt{>}):}
\begin{quote}
\texttt{<}
\end{quote}
}
\end{examplebox}
\newpage
\subsubsection{Appending completion sentence.}
\label{app:appending_sentence}
\begin{examplebox}{Prompt: Completion Sentence Generation}
{\small
\textbf{Instruction:} \\
You are given a Wikipedia passage that contains an entity. Then you are given the entity’s name. Produce one additional sentence that naturally appends to the passage, reaffirming the entity’s identity. You may use one of the following formats (or a similarly natural variant):
\begin{itemize}
  \item “The \textit{[description]} is called \textit{[entity]}.”
  \item “The \textit{[description]} is named \textit{[entity]}.”
  \item “Locals called it \textit{[entity]}.”
  \item “They refer to it as \textit{[entity]}.”
\end{itemize}

\textbf{IMPORTANT:}
\begin{enumerate}
  \item You must use the \textbf{exact entity name as provided}—no alterations, changes in capitalization, or partial usage.
  \item Your output should only be \textbf{that single appended sentence}.
  \item The sentence should be generic and do a good job thoroughly introducing the entity. It should naturally lead up to naming the entity, so that if the entity were removed, a model would be likely to complete the sentence with that entity.
\end{enumerate}

\textbf{One-Shot Example}

\textbf{Passage:}
\begin{quote}
\texttt{< Frankenstein is a gothic novel by Mary Shelley that was first published in 1818. The story follows a young scientist who\\ creates a sapient creature through an unorthodox experiment, and it is often hailed as the first true work of science fiction. >}
\end{quote}

\textbf{Entity:}
\begin{quote}
\texttt{< Mary Shelley >}
\end{quote}

\textbf{Output:}
\begin{quote}
\texttt{< The author of the novel Frankenstein is named Mary Shelley. >}
\end{quote}

\textit{Notice how:}
\begin{itemize}
  \item The entity is introduced \textbf{exactly as given}.
  \item The sentence flows \textbf{naturally from the passage}.
\end{itemize}

\vspace{1em}

\textbf{Now it is your turn.}

\textbf{Entity:}
\begin{quote}
\texttt{< \{entity\} >}
\end{quote}

\textbf{Original Passage:}
\begin{quote}
\texttt{< \{passage\} >}
\end{quote}

\textbf{Output (ending with \texttt{>}):}
\begin{quote}
\texttt{<}
\end{quote}
}
\end{examplebox}
\newpage
\subsection{SQuAD and WebQuestions}
\label{app:oneshot-prompts}
One-shot prompts for SQuAD and WebQ.
\begin{examplebox}{SQuAD}
{\small
Context: Frank Herbert was an American science‑fiction author best known for his novel Dune.
Example Q: Who wrote Dune? A: Frank Herbert\\
Context: ...\\
Q: ... A: ...}
\end{examplebox}
\begin{examplebox}{WebQuestions}
{\small
Example Q: Who wrote “Dune”? A: Frank Herbert \\
Q: ... A: ...}
\end{examplebox}

\end{document}